%% file: draftmain.tex
\documentclass{article}
\usepackage{iclr2026/iclr2026_conference,times}
\input{iclr2026/math_commands.tex}

\usepackage[T1]{fontenc}
\usepackage[utf8]{inputenc}
\usepackage{microtype}
\usepackage{mathtools}

\usepackage{booktabs}
\usepackage{tabularx}
\usepackage{array}
\usepackage{multirow}
\usepackage{makecell}
\usepackage{threeparttable}
\usepackage{siunitx}
\usepackage{graphicx}
\usepackage[export]{adjustbox}
\usepackage[table]{xcolor}
\usepackage{tikz}
\usetikzlibrary{arrows.meta,positioning,calc}
\usepackage[most]{tcolorbox}
\usepackage{float}
\usepackage[section]{placeins}
\usepackage{wrapfig}
\usepackage{subcaption}

\usepackage[font=small,labelfont=bf]{caption}

\usepackage{algorithm}
\usepackage{algpseudocode}
\makeatletter
\providecommand{\theHALG@line}{\thealgorithm.\arabic{ALG@line}}
\makeatother

\usepackage{enumitem}
\usepackage{pifont}
\newcommand{\cmark}{\ding{51}}
\newcommand{\xmark}{\ding{55}}

\usepackage{url}
\usepackage{xurl}
\definecolor{iclrblue}{rgb}{0.0,0.47,0.74}
\usepackage{hyperref}
\hypersetup{
  colorlinks=true,
  linkcolor=iclrblue,
  citecolor=iclrblue,
  urlcolor=iclrblue,
  breaklinks,
  pdfencoding=auto,
  pdftitle={Residual Drift Dominates Contradiction in Multi-Turn Constraint Reasoning},
  pdfauthor={Sebastien Kawada},
  pdfsubject={ICLR 2026 Workshop on Reasoning and Planning for LLMs},
  pdfkeywords={multi-turn reasoning, constraint satisfaction, satisfiable drift, MUS repair, minimal unsatisfiable subsets, constraint ledger, LLM evaluation, DRIFT-Bench, Z3},
}
\usepackage[capitalize,noabbrev]{cleveref}

\definecolor{backred}{RGB}{255, 190, 190}
\definecolor{backblue}{RGB}{210, 230, 250}
\definecolor{backgrey}{RGB}{220, 220, 220}
\definecolor{backgreen}{RGB}{200, 235, 200}
\newcommand{\best}{\cellcolor{backred}}

\newcommand{\gain}{\cellcolor{backgreen}}
\definecolor{shadecolor}{RGB}{237,237,237}

\definecolor{clrboxcolor}{RGB}{245,245,245}
\definecolor{clrframe}{RGB}{0,71,119}
\newtcolorbox{clrbanner}{
  colback=clrboxcolor, colframe=clrframe, boxrule=1pt,
  left=2pt, right=2pt, top=2pt, bottom=2pt,
}

\renewcommand{\arraystretch}{1.0}
\usepackage{xspace}
\newcommand{\Direct}{\textsc{Direct}\xspace}
\newcommand{\CoT}{\textsc{Chain-of-Thought}\xspace}
\newcommand{\LedgerMethod}{\textsc{Ledger}\xspace}
\newcommand{\MUSRepair}{\textsc{MUS-Repair}\xspace}
\newcommand{\DRIFT}{\textsc{DRIFT-Bench}\xspace}

\title{Residual Drift Dominates Contradiction\\in Multi-Turn Constraint Reasoning}

\author{Sebastien Kawada \\
Kaons (\href{https://www.kaons.com}{kaons.com}) \\
Los Angeles, United States \\
\texttt{sebastien@kaons.com}}

\iclrfinalcopy
\begin{document}
\maketitle

%=============================================================================
% ABSTRACT
%=============================================================================
\begin{abstract}
How do multi-turn reasoning systems fail? The expected answer is logical contradiction, in which the system's maintained state becomes unsatisfiable. We show that the dominant mode is instead \emph{satisfiable drift}, where the internal state stays consistent while the returned answer silently violates prior commitments.
We build \DRIFT (Decomposing Reasoning Into Failure Types), a solver-instrumented benchmark of 816 test problems across three constraint domains, and evaluate four methods on it across four open-weight models (8B--120B parameters). MUS-Repair, which feeds minimal unsatisfiable subsets back to the generator, is strongest in every setting ($+1.8$ to $+15.0$ pp over the best non-MUS baseline). But the central finding is what repair leaves behind. After structured feedback, models rarely contradict themselves. They forget. Residual errors are 98--100\% satisfiable drift across all settings, while contradiction drops to near zero. Reliable multi-turn systems must separately validate that the returned answer respects the maintained state. Code is available at \url{https://github.com/kaons-research/drift-bench}.
\end{abstract}

%=============================================================================
% 1  INTRODUCTION
%=============================================================================
\section{Introduction}

When an interactive assistant manages evolving structured state, it must honor every commitment it has already accepted while folding in new constraints. A scheduling tool that confirms ``Bob is not on Tuesday'' should never subsequently place Bob on Tuesday, yet current language models do exactly this with troubling regularity. What makes the failure especially dangerous is its subtlety. The system's internal state remains logically consistent, no solver alarm fires, and the returned answer looks correct to every automated check that inspects only state consistency. We call this pattern \emph{satisfiable drift}, and show that it accounts for the vast majority of residual errors even after structured repair feedback. Figure~\ref{fig:teaser} decomposes residual errors by channel: drift dominates across every model, while contradiction is near-invisible (Table~\ref{tab:intro-summary}).

\begin{figure}[!ht]
\centering
\begin{minipage}[b]{0.46\textwidth}
    \centering
    % Drift/Contradiction/Other decomposition — TikZ horizontal stacked bars
    \definecolor{driftbar}{RGB}{65,150,95}    % muted green
    \definecolor{contrabar}{RGB}{190,70,70}   % muted red
    \definecolor{otherbar}{RGB}{200,200,195}  % warm gray
    \pgfmathsetmacro{\bW}{4.8}% total bar width in cm
    \resizebox{\linewidth}{!}{\begin{tikzpicture}[
        y=0.78cm,
        lbl/.style={font=\scriptsize\bfseries, text=white, anchor=center},
        mlbl/.style={font=\footnotesize, anchor=east},
    ]
    \newcommand{\driftrow}[7]{%
      % #1=row index, #2=model label, #3=drift%, #4=contra%, #5=other%, #6=contraLabel, #7=barHeight
      \pgfmathsetmacro{\ylo}{#1*0.82}
      \pgfmathsetmacro{\yhi}{\ylo+#7}
      \pgfmathsetmacro{\ymid}{\ylo+#7/2}
      \pgfmathsetmacro{\dw}{\bW*#3/100}
      \pgfmathsetmacro{\cw}{\bW*#4/100}
      \pgfmathsetmacro{\ow}{\bW*#5/100}
      \pgfmathsetmacro{\dend}{\dw}
      \pgfmathsetmacro{\cend}{\dw+\cw}
      \pgfmathsetmacro{\dmid}{\dw/2}
      \pgfmathsetmacro{\omid}{\dw+\cw+\ow/2}
      % Model label
      \node[mlbl] at (-0.08, \ymid) {#2};
      % Bars
      \fill[driftbar, rounded corners=1pt]  (0, \ylo) rectangle (\dend, \yhi);
      \fill[contrabar] (\dend, \ylo) rectangle (\cend, \yhi);
      \fill[otherbar, rounded corners=1pt]  (\cend, \ylo) rectangle (\bW, \yhi);
      % Drift percentage label (inside bar)
      \node[lbl] at (\dmid, \ymid) {#3\%};
      % Other % is intentionally not labeled inside the bar — the gray width conveys the
      % share and Table 6 reports the exact number. Avoids clipping on narrow "Other" bars.
      % Contradiction label (right of bar, red)
      \node[font=\tiny, text=contrabar, anchor=west] at (\bW+0.1, \ymid) {#6};
    }

    \driftrow{0}{Qwen3-8B}{100.0}{0.0}{0.0}{0.0\%}{0.46}
    \driftrow{1}{Qwen3-32B}{98.1}{1.9}{0.0}{1.9\%}{0.46}
    \driftrow{2}{gpt-oss-20b}{99.9}{0.1}{0.0}{0.1\%}{0.46}
    \driftrow{3}{gpt-oss-120b}{99.9}{0.1}{0.0}{0.1\%}{0.46}

    % Legend
    \fill[driftbar, rounded corners=1pt]  (0, -0.42) rectangle (0.24, -0.24);
    \node[font=\tiny, anchor=west] at (0.30, -0.33) {Drift};
    \fill[contrabar, rounded corners=1pt] (1.0, -0.42) rectangle (1.24, -0.24);
    \node[font=\tiny, anchor=west] at (1.30, -0.33) {Contradiction};
    \fill[otherbar, rounded corners=1pt]  (2.75, -0.42) rectangle (2.99, -0.24);
    \node[font=\tiny, anchor=west] at (3.05, -0.33) {Other};

    % Force bounding box to include all labels
    \useasboundingbox (-1.75, -0.6) rectangle (\bW+0.75, 3.15);

    \end{tikzpicture}}
    \caption{Residual error decomposition after \MUSRepair. Drift (answer violates a SAT ledger) accounts for 98--100\% of residual errors; contradiction (red, at right) is near-invisible.}
    \label{fig:teaser}
\end{minipage}%
\hfill
\begin{minipage}[b]{0.52\textwidth}
    \centering
    \captionsetup{type=table}% make the next \caption register as a table, not a figure
    \footnotesize
    \setlength{\tabcolsep}{2.8pt}
    \renewcommand{\arraystretch}{0.92}
    \resizebox{\linewidth}{!}{%
    \begin{tabular}{@{}l rr rr r@{}}
        \toprule
        & \multicolumn{2}{c}{\textbf{Best baseline}} & \multicolumn{2}{c}{\textbf{\MUSRepair}} & \\
        \cmidrule(lr){2-3}\cmidrule(lr){4-5}
        \textbf{Model} & {Acc.} & {Method} & {Acc.} & {Drift\,\%} & {$\Delta$\,(pp)} \\
        \midrule
        Qwen3-8B  & 28.2 & Direct & \best 30.0 & 100.0 & \gain +1.8 \\
        Qwen3-32B & 31.4 & CoT    & \best 38.2 & 98.1 & \gain +6.8 \\
        gpt-oss-20b & 53.7 & Ledger & \best 68.7 & 99.9 & \gain +15.0 \\
        gpt-oss-120b & 54.0 & CoT  & \best 62.7 & 99.9 & \gain +8.7 \\
        \bottomrule
    \end{tabular}%
    }
    \caption{Summary of main results. \MUSRepair outperforms the strongest non-MUS baseline in every setting. Drift\,\% shows the share of residual errors from satisfiable drift rather than contradiction.}
    \label{tab:intro-summary}
\end{minipage}
\end{figure}

Existing evaluations collapse two fundamentally different failure modes into a single accuracy number \citep{wei2022chain,yao2024collie,madaan2023selfrefine}. \emph{Contradiction}, where the maintained state becomes unsatisfiable, is a state-level defect that formal methods can detect. \emph{Satisfiable drift}, where the state is consistent but the assignment violates it, requires a second verification layer most systems lack. This paper separates the two with a solver-instrumented benchmark that checks both ledger satisfiability and assignment validity at every turn across 816 problems and four open-weight models (Table~\ref{tab:intro-summary}).

\textbf{Findings.}
\ding{182}~\MUSRepair is the strongest method in every setting, producing gains of $+1.8$ to $+15.0$~pp over the best non-MUS baseline, all of which survive paired tests after false-discovery correction.
\ding{183}~These gains do not eliminate the dominant failure mode. After structured feedback, 98--100\% of remaining failures involve a consistent ledger with a violating assignment, while contradiction drops to near zero. Models stop contradicting themselves but keep forgetting prior commitments.
\ding{184}~The degradation with conversational depth is structural rather than a capacity bottleneck. Even gpt-oss-120b drops from 93\% at turn one to 40\% at turn ten; higher capability lifts the entire curve but does not flatten it.

\textbf{Contributions.}
\ding{182}~\textbf{\DRIFT}, a solver-instrumented multi-turn benchmark covering three constraint domains (logic grid, scheduling, seating) with Z3-verified turn-level decomposition of contradiction and drift.
\ding{183}~A \textbf{trigger-conditioned repair interface} that routes unsatisfiable states through MUS localization and satisfiable assignment failures through policy diagnostics within a single retry loop.
\ding{184}~The \textbf{first empirical demonstration} that satisfiable drift dominates residual errors across all tested settings, arguing that contradiction and drift should be reported as separate evaluation metrics.

%=============================================================================
% 2  RELATED WORK
%=============================================================================
\section{Related Work}

\paragraph{Evaluation of multi-step reasoning.}
Prompting strategies, search over intermediate traces, and tool-augmented agent architectures have produced substantial accuracy gains on reasoning benchmarks \citep{wei2022chain,kojima2022large,wang2023selfconsistency,yao2023tree,gao2023pal,chen2023program,yao2023react,hu2024hiagent,han2025verifiagent}. These advances primarily target single-turn performance or final-answer quality, and they have achieved impressive results in that scope. However, most evaluations do not instrument turn-level state validity under accumulated constraints. The COLLIE benchmark \citep{yao2024collie} evaluates LLMs on constraint satisfaction, but it operates in a single-turn setting without multi-turn state tracking or failure-channel decomposition. Long-context and length-extrapolation studies document sensitivity to sequence length and position \citep{press2022trainshort,liu2024lost}, yet they do not separate state inconsistency from assignment inconsistency conditional on a satisfiable state. Our benchmark is designed to fill this gap. Each turn is solver-verified for both ledger satisfiability and assignment validity.

\paragraph{Verifier-guided repair and self-correction.}
Iterative self-correction with verifier feedback produces strong aggregate improvements in mathematics and code \citep{cobbe2021training,lightman2023let,madaan2023selfrefine,shinn2023reflexion}. Tool-integrated reasoning systems, including systems that couple deterministic solvers with neural generation \citep{lyu2023faithful,lu2024chameleon}, improve single-turn accuracy, with \citet{lyu2023faithful} demonstrating that removing the deterministic external solver causes a 50-point accuracy drop on GSM8K. However, aggregate gains can obscure shifts in the composition of residual errors. Endpoint accuracy may improve substantially even as assignment-level drift remains unchanged or worsens, because the error types that repair eliminates are not necessarily the ones most visible to users. Recent work on the limits of LLM self-verification reaches a related conclusion. \citet{stechly2025limitsverify} show that when GPT-4 is tasked with both generating and critiquing its own answers, performance actually decreases, and that substantial gains require a sound external verifier regardless of critique richness. Our analysis extends this concern to interactive trajectories by decomposing residuals by operational failure type.

\paragraph{Formal methods in neural systems.}
Satisfiability solving and minimal unsatisfiable subset extraction are well-established tools in symbolic debugging and verification \citep{deMoura2008z3,liffiton2008algorithms,belov2012muser2,biere2009handbook}. A separate but related thread comes from task-oriented dialogue, where belief state updates track evolving user requirements \citep{young2013pomdp,wu2019transferable}. The ledger mechanism in our system draws on both traditions. It maintains formal constraint sets, as in symbolic verification, but updates them incrementally at each conversational turn, as in dialogue state tracking. Our contribution is adapting this combined toolbox to neural multi-turn traces through fixed turn-level solver instrumentation, trigger-conditioned repair routing, and paired inferential analysis over interactive trajectories.

%=============================================================================
% 3  METHOD
%=============================================================================
\section{Method}
\label{sec:method}

\subsection{Notation and State Semantics}

The multi-turn setting requires distinguishing between the raw model output and the structured state derived from it. We write \(u_t\) for the user message at turn \(t\), \(a_t\) for the model's response text, and \(A_t\) for the structured assignment parsed from \(a_t\) when parsing succeeds. The cumulative gold constraints are denoted by \(\mathcal{C}_{1:t}\), extracted constraints by \(\widehat{\mathcal{C}}_t\) (the model's parse of new constraints at turn \(t\)), and ledger state by \(L_t\). The predicate \(\mathrm{SAT}(\cdot)\) indicates solver satisfiability; we write \(\mathrm{UNSAT}(\cdot)\) for its negation.

Each problem is a turn sequence $\{u_t\}_{t=1}^{T}$ with cumulative gold constraints

\[
\mathcal{C}_{1:t} = \bigcup_{\tau=1}^{t}\mathcal{C}_{\tau}^{\text{new}}.
\]
Turn-level correctness is defined by constraint satisfaction rather than string match against a single witness assignment. The operational correctness predicate applies to the raw response and its parsed assignment:

\[
\mathrm{Correct}(a_t) = \mathrm{Parse}(a_t)\ \land\ \mathrm{Complete}(A_t)\ \land\ \mathrm{Satisfies}(A_t,\mathcal{C}_{1:t}).
\]

In implementation, \texttt{answer\_correct} is obtained by checking satisfiability of $\mathcal{C}_{1:t}$ with the parsed $A_t$ injected as an assignment in Z3. This definition remains valid when multiple satisfying assignments exist.
The same \(\mathrm{Satisfies}\) predicate appears in drift diagnostics with $L_t$ replacing $\mathcal{C}_{1:t}$. The constraint set argument determines which notion of consistency is tested.
We measure accuracy against gold cumulative constraints \(\mathcal{C}_{1:t}\), while drift is a diagnostic defined against the model-maintained ledger \(L_t\).

The distinction between ledger satisfiability and assignment validity is central to the paper. A turn can preserve $\mathrm{SAT}(L_t)$ while still violating active commitments through $\neg \mathrm{Satisfies}(A_t,L_t)$. This separation allows contradiction and drift to be measured as distinct channels rather than merged into a single error indicator.
Formally, let $\Phi(A_t)$ denote the assignment constraints induced by the parsed answer. Then

\[
\mathrm{Satisfies}(A_t,S)=\mathrm{SAT}(S\cup \Phi(A_t)).
\]

The parser predicate \(\mathrm{Parse}(a_t)\) is one only when the response is valid schema-conforming JSON for the domain. The completeness predicate \(\mathrm{Complete}(A_t)\) is one only when each required entity is assigned exactly once. The ledger update is

\[
\mathrm{Merge}(L_{t-1},\widehat{\mathcal{C}}_t)=L_{t-1}\cup \mathrm{Dedup}(\widehat{\mathcal{C}}_t),
\]

where \(\mathrm{Dedup}\) removes canonical duplicates before insertion.

These predicates partition turn outcomes into three categories. A turn is \emph{consistent} when the ledger is satisfiable and the assignment respects it. When the ledger remains satisfiable but the assignment violates it, the turn exhibits \emph{drift}. When the ledger itself becomes unsatisfiable, the turn exhibits \emph{contradiction}. The critical distinction is that drift produces no solver alarm, making it invisible to any system that checks only state consistency. Figure~\ref{fig:method-comparison} illustrates all three outcomes on a four-turn scheduling trajectory where drift occurs at the final turn.

\subsection{System Components}

The evaluation system decomposes each turn into four stages, reflecting a deliberate separation of generation from verification. A generator $G$ produces the response $a_t$ given the current user message and prior ledger state. An extractor $E$ then parses the response alongside the user message to identify newly introduced constraints $\widehat{\mathcal{C}}_t$. These feed into a verifier $V$, which runs both solver-level satisfiability checks and policy-level checks on the parsed assignment. Finally, a repair policy $R$ examines the verifier output and decides whether to issue a retry with targeted feedback.

\begin{wrapfigure}{r}{0.52\linewidth}
\vspace{-\intextsep}
\begin{minipage}{\linewidth}
\begin{algorithm}[H]
    \caption{Turn processing with verification and optional repair.}
    \label{alg:turn-proc}
    \scriptsize
    \begin{algorithmic}[1]
        \Statex \textbf{Input:} $u_t$, $L_{t-1}$, method $m$, turn $t$, repair budget $k$
        \Statex \textbf{Output:} response $a'_t$, ledger $L_t$
        \State $a_t \gets G(u_t, L_{t-1})$
        \State $\widehat{\mathcal{C}}_t \gets E(u_t, a_t, t)$
        \State $L_t \gets L_{t-1} \cup \mathrm{Dedup}(\widehat{\mathcal{C}}_t)$
        \State $(\mathrm{sat}_t,\,\mathcal{T}_t) \gets V(L_t, a_t)$
        \If{$m \neq \textsc{MUS-Repair}$ \textbf{or} $(\mathrm{sat}_t \,\land\, \mathcal{T}_t\!=\!\emptyset)$}
            \State \Return $(a_t, L_t)$
        \EndIf
        \State $a'_t \gets a_t$
        \For{$i = 1$ \textbf{to} $k$}
            \State $\mathcal{U}_t \gets \mathrm{MUS}(L_t)$ \textbf{if} $\neg\,\mathrm{sat}_t$ \textbf{else} $\emptyset$
            \State $a'_t \gets R\bigl(u_t, L_{t-1}, \mathrm{Render}(\mathcal{T}_t, \mathcal{U}_t)\bigr)$
            \State $L_t \gets L_{t-1} \cup \mathrm{Dedup}\bigl(E(u_t, a'_t, t)\bigr)$
            \State $(\mathrm{sat}_t,\,\mathcal{T}_t) \gets V(L_t, a'_t)$
            \If{$\mathrm{sat}_t \,\land\, \mathcal{T}_t\!=\!\emptyset$} \textbf{break} \EndIf
        \EndFor
        \State \Return $(a'_t, L_t)$
    \end{algorithmic}
\end{algorithm}
\end{minipage}
\vspace{-\intextsep}
\end{wrapfigure}

The verifier combines solver-level satisfiability checks with policy-level checks on the parsed assignment, emitting a deterministic trigger code per failure type. The five codes are \textit{Answer-Ledger Conflict} (ledger is SAT but the assignment violates it), \textit{Unsatisfiable Ledger} (ledger is UNSAT), \textit{Incomplete Assignment} (required entities missing), \textit{Answer Parse Failure} (invalid JSON), and \textit{Constraint Extraction Failure} (no constraints extracted). At runtime these codes route the repair decision; post-hoc they enable fine-grained failure decomposition. Algorithm~\ref{alg:turn-proc} shows how they integrate into the turn processing loop.

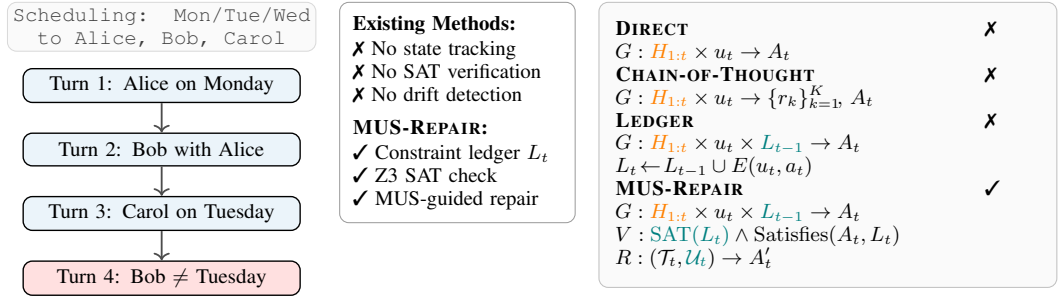
\begin{figure*}[!t]
\centering
\resizebox{\textwidth}{!}{\input{figures/method_comparison_v2.tex}}\\[4pt]
\caption{\textbf{Comparison of constraint reasoning approaches.} Left: a four-turn scheduling trajectory where drift occurs at turn~4 (red). Center: properties of baselines vs.\ \MUSRepair. Right: formal method signatures with \textcolor{orange}{orange} marking implicit context accumulation and \textcolor{teal}{teal} marking explicit ledger state and solver verification. Only \MUSRepair detects the drift because it verifies both $\mathrm{SAT}(L_t)$ and $\mathrm{Satisfies}(A_t, L_t)$.}
\label{fig:method-comparison}
\end{figure*}

We evaluate four inference policies on this shared infrastructure: \Direct, \CoT, \LedgerMethod, and \MUSRepair. Figure~\ref{fig:method-comparison} compares their formal signatures on a scheduling example where drift occurs at the final turn. The repair step is active only for \MUSRepair and only when the verifier emits one or more failing triggers. Crucially, we hold the extractor and verifier fixed across all methods so that observed differences reflect reasoning and repair strategy, not variation in parsing or verification logic. This shared-infrastructure design isolates the comparison to the reasoning policy itself.

\subsection{Minimal Unsatisfiable Subset for Repair}

When $V$ detects an unsatisfiable ledger state, \MUSRepair computes a minimal unsatisfiable subset $\mathcal{U}_t \subseteq L_t$ such that
\[
\mathrm{UNSAT}(\mathcal{U}_t)=1,\quad
\forall \mathcal{U}' \subset \mathcal{U}_t,\ \mathrm{SAT}(\mathcal{U}')=1.
\]
This subset is minimal in set inclusion and identifies a minimal committed constraint subset that is jointly inconsistent at turn $t$. The retry prompt receives trigger diagnostics and, for unsatisfiable states, the corresponding $\mathcal{U}_t$. The same retry channel is used for satisfiable assignment failures through policy triggers, so contradiction and drift are handled in one controlled repair interface. The repair feedback packet is
\[
F_t =
\begin{cases}
(\mathcal{T}_t,\mathcal{U}_t), & \mathrm{UNSAT}(L_t)=1,\\
(\mathcal{T}_t,\emptyset), & \mathrm{SAT}(L_t)=1\ \land\ \mathcal{T}_t\neq\emptyset.
\end{cases}
\]
MUS is injected only for contradiction events, while satisfiable assignment failures are repaired with policy diagnostics and the prior ledger state.

\subsection{Failure Channel Decomposition}

The conceptual contradiction indicator is
$I^{\text{unsat}}_t = \mathbf{1}[\mathrm{UNSAT}(L_t)].$
The conceptual drift indicator is
$I^{\text{drift}}_t = \mathbf{1}[\mathrm{SAT}(L_t)\land \neg \mathrm{Satisfies}(A_t,L_t)].$
Current logs provide direct contradiction status and trigger-level diagnostics on \MUSRepair traces. We measure contradiction with \texttt{z3\_sat}=0 and drift with the Answer-Ledger Conflict trigger \texttt{answer\_ledger\_conflict}, which corresponds to a satisfiable ledger with a violating assignment. Parser and completeness failures are tracked by Answer Parse Failure, Incomplete Assignment, and Constraint Extraction Failure triggers.

Primary reporting uses turn-level accuracy as defined by \(\mathrm{Correct}(a_t)\). The inference protocol is described in Section~\ref{sec:inference-protocol}.

%=============================================================================
% 4  EXPERIMENTAL SETUP
%=============================================================================
\section{Experimental Setup}

\subsection{\DRIFT}

\DRIFT problems are generated by a procedure that guarantees every gold interaction trajectory is satisfiable at each turn. For each domain $\mathcal{D}\in\{\texttt{logic\_grid},\texttt{scheduling},\texttt{seating}\}$, the generator samples entities, contextual framing, and one to three candidate constraints per turn. It accepts a candidate set only when the cumulative set remains satisfiable under Z3

\[
\mathrm{SAT}\!\left(\mathcal{C}_{1:t-1}\cup \widehat{\mathcal{C}}^{\text{cand}}_t\right)=1.
\]

If the candidate set is unsatisfiable, the turn is resampled until acceptance or retry budget exhaustion. This process ensures that every gold interaction trajectory is satisfiable at each turn.
Generation also removes duplicate constraints by canonical form before satisfiability checks, which prevents trivial repetition across turns.

Gold correctness does not assume a unique assignment. We verify by checking satisfiability of cumulative constraints conjoined with the parsed answer assignment, so a response is correct whenever it satisfies the active constraints, even if multiple assignments are valid.

Each domain uses a fixed template that determines the structural parameters of generated problems. Logic-grid instances pair four entities with three categorical attributes, producing compact but combinatorially rich assignments. Scheduling instances involve five to seven events assigned to temporal slots, with predicates such as ordering and simultaneity constraints. Seating instances are the most spatially complex, placing six to eight participants around round or rectangular tables subject to adjacency, separation, and positional constraints. Turn count is sampled between four and ten, with one to three new constraints introduced per turn.

The final corpus has 1,020 problems with a fixed seed split of 816 test and 204 development instances. Table~\ref{tab:benchmark-stats} summarizes structural properties by domain; the Final column is the mean number of cumulative active constraints at the last turn.

\begin{table}[htb]
    \centering
    \caption{\DRIFT structure by domain.}
    \label{tab:benchmark-stats}
    \small
    \input{tables/benchmark_stats}
\end{table}

\subsection{Evaluation Stack and Model Matrix}

All methods run in a shared OpenAI-compatible serving stack with identical extraction, verification, and logging paths. The model matrix $\mathcal{M}$ contains Qwen3-8B, Qwen3-32B, gpt-oss-20b, and gpt-oss-120b, the method set is $\Pi=\{\Direct,\CoT,\LedgerMethod,\MUSRepair\}$, and $\mathcal{P}$ is the 816-problem test split. The Qwen models are from the Qwen3 family \citep{qwen3-2025}. The gpt-oss models are OpenAI's open-weight releases at 20B and 120B parameters \citep{openai2025gptoss}, served locally through vLLM \citep{kwon2023vllm} under the same stack as the Qwen runs. For gpt-oss evaluations we use deterministic decoding with temperature set to zero and default reasoning configuration, with paired comparisons run under fixed decoding controls within each model.

The full corpus sums to $|\mathcal{M}|\cdot|\Pi|\cdot\sum_{p\in\mathcal{P}} T_p = 4\times 4\times 5{,}672 = 90{,}752$ turn evaluations. The gpt-oss-120b run is complete for all methods at 5,672 rows and 816 problems per method and is included in all main-text tables.

\subsection{Inference Protocol and Robustness Checks}
\label{sec:inference-protocol}

To ensure that reported accuracy differences are not artifacts of problem sampling, we construct 95\% bootstrap confidence intervals at the problem level and assess pairwise significance using sign-permutation tests against \Direct. We then apply Benjamini-Hochberg correction across all reported comparisons to control the false-discovery rate.

Prompt templates, JSON schema constraints, repair message format, and extraction prompts are documented in Appendix~A. Runtime controls are fixed across methods with temperature set to zero, maximum repair attempts set to two, maximum truncation retries set to two, and ledger token budget set to 3,000 unless a serving-side safety clamp is required.

One practical consideration is that the gpt-oss models occasionally produce truncated responses, which could in principle affect accuracy estimates. We verified that restricting analysis to non-truncated responses preserves both the method ordering and the MUS-Repair margins, indicating that truncation is not a systematic confound.

%=============================================================================
% 5  RESULTS
%=============================================================================
\section{Results}

\subsection{Primary Accuracy}

Table~\ref{tab:inferential} presents the full results. \textbf{\MUSRepair is the strongest method in every model setting}, with gains over \Direct ranging from $+2.0$~pp on Qwen3-8B to $+16.2$~pp on gpt-oss-20b. Every MUS-Repair comparison survives paired problem-level permutation tests after Benjamini-Hochberg correction ($q_\mathrm{FDR} < 0.03$ in all cases), and the pattern holds when tested against each model's strongest non-MUS comparator rather than \Direct alone (Appendix Table~\ref{tab:app-inferential-best-nonmus}). Structured repair helps the weakest model (Qwen3-8B, 30\%) and the strongest (gpt-oss-20b, 69\%) alike. This consistency across a wide capability range suggests that the benefit comes from the verification-and-retry mechanism itself rather than from model-specific artifacts. The other methods show mixed results. Ledger significantly hurts Qwen3-8B ($-3.1$~pp, $q = 0.003$) and gpt-oss-120b ($-2.3$~pp, $q = 0.022$), while CoT produces modest gains on Qwen3-32B and gpt-oss-120b but not on the other two models.

\begin{table}[!t]
    \centering
    \caption{Turn-level accuracy, paired inferential tests versus \Direct ($n=816$), and depth retention. \colorbox{iclrblue!8}{Highlighted} rows show \MUSRepair. Retain\,\% $=$ turn-10 accuracy $/$ turn-1 accuracy, measuring how well each method preserves performance as constraints accumulate.}
    \label{tab:inferential}
    \footnotesize
    \setlength{\tabcolsep}{3pt}
    \renewcommand{\arraystretch}{0.92}
    \resizebox{\linewidth}{!}{\input{tables/inferential_tests}}
\end{table}

\subsection{Capability Scaling of Repair Gains}

\textbf{Stronger models benefit more from structured repair.} Raw accuracy gains are larger for more capable models, but comparing absolute improvements across models with different baselines can be misleading. A more informative measure is relative lift, which normalizes the \MUSRepair gain by each model's best non-MUS baseline.
\[
\rho_m=\frac{A_m^{\text{MUS}}-\max\!\left(A_m^{\text{Direct}},A_m^{\text{CoT}},A_m^{\text{Ledger}}\right)}
{\max\!\left(A_m^{\text{Direct}},A_m^{\text{CoT}},A_m^{\text{Ledger}}\right)}.
\]
Relative lift rises from 6.4\% on Qwen3-8B to 27.9\% on gpt-oss-20b before dropping to 16.2\% on gpt-oss-120b. The non-monotonic drop at gpt-oss-120b resists simple scaling predictions, but the overall trajectory still shows that repair remains materially beneficial even at the highest capability level tested. One plausible explanation centers on instruction-following fidelity. The repair signal is a structured prompt containing trigger codes and a minimal unsatisfiable subset. Converting this signal into a corrected assignment requires precisely the kind of structured instruction following that improves with model capability. A model that cannot parse the signal treats the retry as noise.

Ledger-only tracking, by contrast, is not uniformly positive. It hurts Qwen3-8B ($-3.0$~pp), is near-neutral on Qwen3-32B, helps gpt-oss-20b ($+1.9$~pp), and drops again on gpt-oss-120b ($-2.1$~pp). Explicit state tracking trades control benefits against the cost of occupying prompt context, and the balance shifts with model capability (Section~\ref{sec:discussion}).

\subsection{Depth Degradation}

\begin{figure}[htb]
    \centering
    \includegraphics[width=0.85\linewidth]{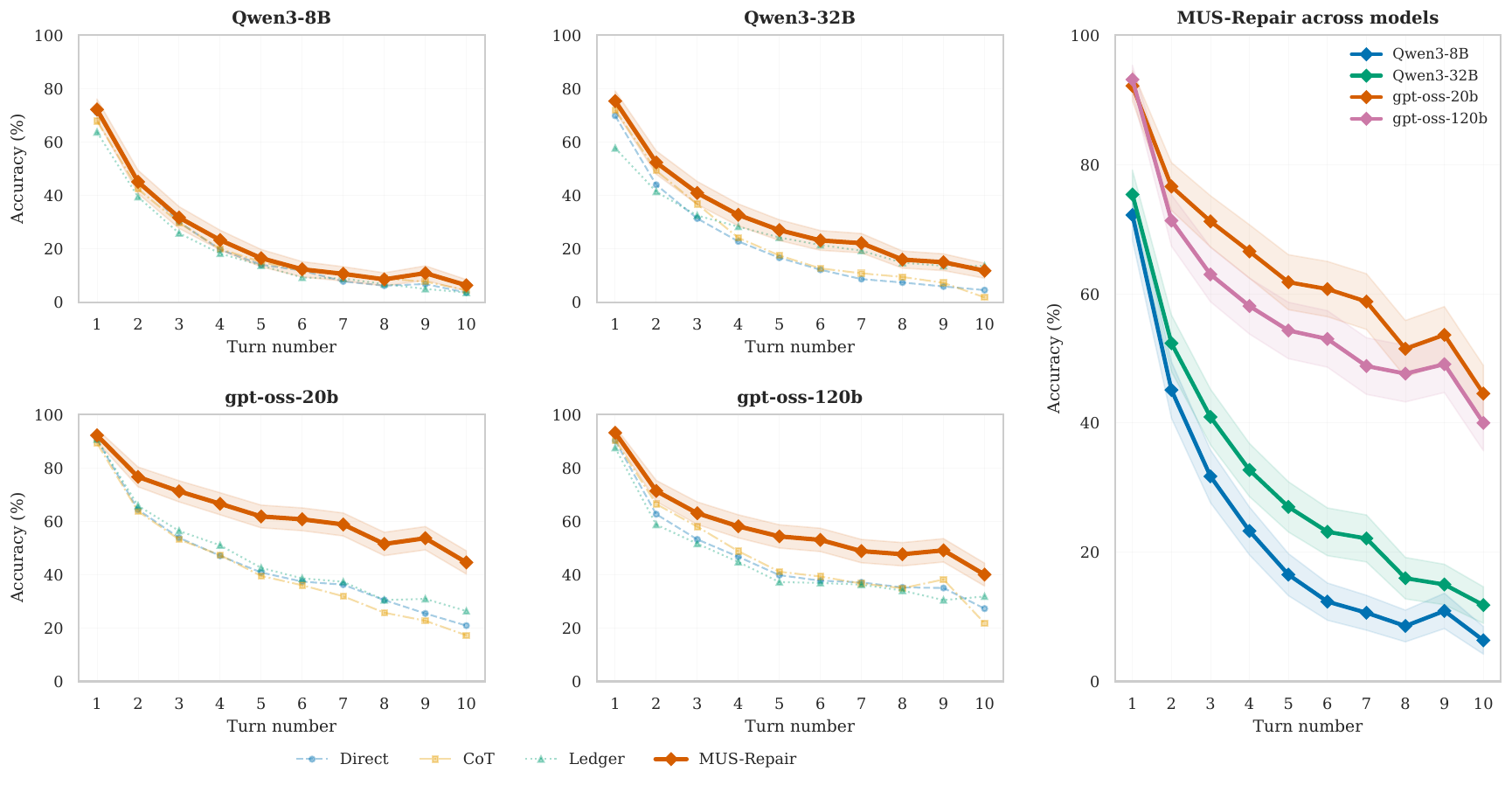}
    \caption{Per-turn accuracy curves. Left $2\times 2$ grid: one panel per model, each showing all four methods (colors encode \emph{method}). Right panel: \MUSRepair across all four models (colors encode \emph{model}). Shaded bands are 95\% bootstrap intervals. Higher capability lifts the curve but does not flatten it.}
    \label{fig:turn-depth-curves}
\end{figure}

\textbf{Every model shows steep accuracy decline with turn depth} (Figure~\ref{fig:turn-depth-curves}). The decline is dramatic across the full model range. Qwen3-8B drops from 72\% at turn one to 6\% at turn ten under \MUSRepair, and even gpt-oss-120b falls from 93\% to 40\% over the same span. Crucially, the shape of the decline is steep rather than gradual, consistent with the probability of violating at least one constraint growing combinatorially as the active set expands. Higher capability lifts the entire curve but does not flatten it. This pattern is reminiscent of the positional sensitivity documented by \citet{liu2024lost} for long contexts, but here the degradation is temporal rather than positional. Long-horizon state maintenance appears to be a qualitatively harder problem that will likely require architectural support beyond pure scaling.

\subsection{Domain Structure}

\textbf{Seating is the hardest domain and scheduling the easiest across all models} (Table~\ref{tab:domain-mus-main}). This ranking holds consistently from the smallest Qwen model to the largest gpt-oss run, indicating that the difficulty is inherent in the constraint topology rather than an artifact of any single model's weaknesses. Seating problems involve circular positional constraints (adjacency, separation, wrapping) that require globally consistent placement of 6--8 entities, whereas scheduling admits more localized solutions. The gap is largest on gpt-oss-20b, where scheduling reaches 85.7\% while seating remains at 38.7\%.

\begin{table}[htb]
    \centering
    \caption{\MUSRepair domain-conditioned accuracy (\%). Scheduling is consistently easiest and seating hardest.}
    \label{tab:domain-mus-main}
    \small
    \begin{tabular}{l S[table-format=2.1] S[table-format=2.1] S[table-format=2.1]}
    \toprule
    Model & {Logic-Grid (\%)} & {Scheduling (\%)} & {Seating (\%)} \\
    \midrule
    Qwen3-8B & 43.1 & 34.7 & 12.1 \\
    Qwen3-32B & 43.9 & 55.7 & 14.8 \\
    gpt-oss-20b & 81.4 & 85.7 & 38.7 \\
    gpt-oss-120b & 64.2 & 87.2 & 36.3 \\
    \bottomrule
    \end{tabular}
\end{table}

%=============================================================================
% 6  FAILURE ANALYSIS
%=============================================================================
\section{Failure Analysis}

\subsection{Contradiction and Drift Decomposition}

Because \MUSRepair logs solver status and trigger codes at every retry, we can decompose residual errors into three distinct channels: contradiction, in which the ledger becomes unsatisfiable; drift, in which the ledger remains satisfiable but the assignment violates it; and formatting or extraction errors. We measure contradiction by \texttt{z3\_sat}=0 and drift by the Answer-Ledger Conflict trigger, which fires when a satisfiable ledger accompanies a returned assignment that violates active constraints.

\textbf{Contradiction repair does not remove the dominant residual failure mode.} This is the paper's central empirical finding. Drift accounts for 98.1\% to 100.0\% of residual errors across all settings (Table~\ref{tab:sat-wrong}), while contradiction drops to 0.0--1.9\%, with near-zero counts in the gpt-oss runs. The magnitude of this imbalance is best appreciated through raw counts. On Qwen3-8B, all 3{,}970 residual errors are drift, with zero contradiction events. On Qwen3-32B, which triggers unsatisfiable states more frequently due to aggressive constraint extraction, contradiction still accounts for only 66 of 3{,}504 residual errors (1.9\%). The remaining errors, categorized as ``Other,'' encompass parse failures, incomplete assignments, and extraction errors. These are comparatively rare and model-dependent. On gpt-oss-20b, Answer Parse Failure triggers fire 664 times and Constraint Extraction Failure triggers fire 559 times across repair retries (Table~\ref{tab:trigger-breakdown-main}), but are absorbed by successful retries before becoming residual errors. Only 1 contradiction and 0 extraction failures remain in the residual set. The practical implication is that reducing unsatisfiable ledgers is necessary for reliability, but most remaining user-facing errors stem from assignments that violate a satisfiable maintained state.

\begin{table}[!htb]
\centering
\caption{\MUSRepair failure channel decomposition over residual errors. The visual decomposition appears in Figure~\ref{fig:teaser}.}
\label{tab:sat-wrong}
\vspace{2pt}
\small
\setlength{\tabcolsep}{4pt}
\input{tables/sat_but_wrong}
\end{table}

\subsection{Trigger Composition, Repair Outcomes, and Residual Overlap}

Trigger event counts (Table~\ref{tab:trigger-breakdown-main}) reinforce the decomposition. Answer-Ledger Conflict dominates every model, firing 12{,}089 times on Qwen3-8B and 5{,}300 times on gpt-oss-20b. Unsatisfiable Ledger, by contrast, concentrates in Qwen3-32B (212 events) and stays scarce elsewhere ($\leq$4 events on the gpt-oss models). Note that Table~\ref{tab:sat-wrong} counts final-row outcomes while Table~\ref{tab:trigger-breakdown-main} counts trigger events across retries, so magnitudes differ by construction.

\textbf{Not all trigger types respond equally to repair.} Schema-completion failures, such as missing entities or malformed JSON, are relatively straightforward to fix with a retry prompt because the error is localized and well-defined. Assignment-level consistency failures, by contrast, require the model to simultaneously satisfy all active constraints while revising its answer, a much harder task. The data bear this out. On Qwen3-8B, post-repair accuracy reaches 65.5\% for Incomplete Assignment triggers but only 4.0\% for Answer-Ledger Conflict triggers. On Qwen3-32B, the same contrast is 69.8\% versus 5.5\%. This recoverability gap explains why drift persists as the dominant failure mode even after multiple repair attempts.

\begin{table}[!htb]
    \centering
    \caption{Trigger event counts in \MUSRepair traces by model. Answer-Ledger Conflict dominates in every setting, while Unsatisfiable Ledger is concentrated in Qwen3-32B.}
    \label{tab:trigger-breakdown-main}
    \small
    \setlength{\tabcolsep}{3pt}
    \renewcommand{\arraystretch}{0.95}
    \input{tables/trigger_breakdown}
\end{table}

\textbf{Residual error overlap across models is high rather than fragmented.} Qwen3-8B covers 95.0\% of gpt-oss-20b residual \MUSRepair errors, and Qwen3-32B covers 92.3\%. This shared residual set points to common hard regions of the benchmark rather than disjoint model-specific failure pockets. The problems that resist repair for one model tend to resist it for all of them, which suggests the difficulty is intrinsic to the constraint structure rather than tied to any particular model's weaknesses.

%=============================================================================
% 7  DISCUSSION
%=============================================================================
\section{Discussion}
\label{sec:discussion}

The central finding is not that MUS-Repair works, but what it leaves behind. After contradiction-aware repair, the residual error mass concentrates in satisfiable drift rather than in unsatisfiable states. This asymmetry has consequences for system design, scaling expectations, and evaluation methodology.

\paragraph{Drift dominance in deployed systems.}
A system that ships satisfiability checks as its primary reliability gate will miss the majority of user-visible failures in this benchmark family. The failure mode is insidious because it evades every standard check. The internal ledger remains consistent, the solver raises no alarm, and the returned answer nonetheless violates a commitment the user already accepted. Unlike contradiction, which at least signals that something has gone wrong, drift produces confident answers that pass every automated check inspecting only state consistency. For scheduling or resource allocation assistants, this means a user who asks ``remind me of the constraints so far'' receives a valid summary while the assignment silently breaks one of those same constraints. Detecting this class of error requires a second verification layer that explicitly checks the assignment against the maintained state. This architectural requirement parallels the finding of \citet{stechly2025limitsverify} that sound external verification is necessary regardless of critique sophistication.

\paragraph{Why stronger models benefit more from symbolic feedback.}
Relative MUS-Repair lift rises from 6.4\% on Qwen3-8B to 27.9\% on gpt-oss-20b. We see two complementary mechanisms at work, both supported by the trigger data. The first is baseline error composition. On Qwen3-8B, Answer-Ledger Conflict accounts for 12{,}089 of 12{,}308 total triggers (98\%), meaning nearly all repair attempts target drift, a failure type that resists retry. On gpt-oss-20b, the trigger mix is more diverse (5{,}300 drift, 1{,}036 incomplete, 664 parse), giving the repair loop a broader surface of recoverable errors. The second mechanism is instruction-following fidelity. The repair signal is a structured prompt containing trigger codes, violated constraints, and a minimal unsatisfiable subset. Converting this signal into a corrected assignment requires the kind of structured instruction following that improves with model capability; a model that cannot parse the signal treats the retry as noise. The trigger data bear this out indirectly. Post-repair accuracy on Answer-Ledger Conflict triggers rises from 4.0\% on Qwen3-8B to 33.3\% on gpt-oss-20b, a factor-of-eight improvement, suggesting that the larger model is better at acting on the structured feedback even for the hardest failure type. The non-monotonic drop at gpt-oss-120b complicates this picture. One possibility is that the largest model's implicit state tracking is already strong enough that the marginal value of explicit MUS feedback diminishes, even though absolute accuracy still improves.

\paragraph{Depth collapse as accumulation.}
The depth curves in Figure~\ref{fig:turn-depth-curves} present perhaps the most challenging finding for scaling-based solutions. The steep decline, rather than gradual erosion, suggests that each new constraint does not simply add a fixed probability of error. Instead, the probability of violating at least one constraint grows combinatorially with the active set, creating a qualitatively harder problem at each successive turn. This framing suggests that flattening the depth curve will require mechanisms that scale sublinearly with constraint count, such as hierarchical state abstractions or incremental re-verification, rather than relying on raw model capacity alone.

\paragraph{Ledger tracking and context competition.}
Explicit ledger injection helps gpt-oss-20b but hurts Qwen3-8B, and the benefit declines again at gpt-oss-120b. The likely mechanism is context competition. Serializing the ledger (up to 3{,}000 tokens) consumes prompt budget that a smaller model needs for reasoning. A larger model absorbs the overhead and uses the explicit state productively. The renewed decline at gpt-oss-120b suggests that above a capability threshold the model's implicit tracking is competitive with explicit injection, so the added context cost outweighs the control benefit.

\paragraph{Toward drift-targeted repair.}
The current repair loop is contradiction-oriented, identifying minimal unsatisfiable subsets and feeding them back. Drift, by contrast, receives only generic policy diagnostics without localizing which constraints are violated or which entities are misplaced. Closing this gap requires localizing the violated constraints.

\[
\mathcal{V}_t = \bigl\{c \in L_t : \neg\mathrm{SAT}\!\bigl(\{c\}\cup\Phi(A_t)\bigr)\bigr\},
\]

which mirrors the MUS definition structurally: MUS localizes a contradictory subset of the ledger, while $\mathcal{V}_t$ identifies ledger constraints violated by the returned assignment.

\paragraph{Beyond solver-structured domains.}
Our benchmark uses formal constraint sets because they enable sound verification, but satisfiable drift is not specific to constraint satisfaction. Any multi-turn system that maintains evolving commitments can exhibit it. A travel-planning assistant might confirm ``no flights on Sunday'' and later propose a Sunday itinerary; a code-editing agent might acknowledge a variable rename and subsequently reference the old name. In these open-domain settings, detecting drift would require extracting implicit constraints from natural language, likely through entailment-based commitment tracking rather than SAT solving. The core diagnostic question remains the same: is the system's internal state valid, and does the output respect that state? We expect drift to dominate in open-domain settings as well, because the underlying cause, forgetting prior commitments while maintaining a coherent narrative, is a property of how language models process sequential context rather than an artifact of the constraint format.

\paragraph{Evaluation implications.}
The distinction between state-level and assignment-level failure extends beyond our benchmark. Dialogue state trackers, collaborative document editors, and iterative code generators all maintain evolving commitments across turns. We suggest that multi-turn evaluations in these domains similarly decompose errors by whether the system's internal state became invalid or whether the output simply failed to respect a valid state. Reporting the two channels separately would prevent the pattern we document here, where progress on one failure type masks stagnation on the other, and would give practitioners a clearer picture of where reliability investments should be directed.

Taken together, these results argue that contradiction detection, though necessary, is not sufficient for reliable multi-turn systems. A second verification layer that checks the returned assignment against the maintained state is needed to catch the dominant failure mode. Reporting contradiction and drift as separate evaluation channels, rather than merging them into a single accuracy number, would give practitioners a clearer picture of where residual risk concentrates.

%=============================================================================
% 8  LIMITATIONS
%=============================================================================
\section{Limitations}

Several scope limitations bear on the generalizability of our findings. The study evaluates four open-weight models from two families but does not include closed-weight frontier systems or specialist fine-tuned variants, either of which might exhibit different drift-to-contradiction ratios. Failure-channel decomposition relies on the solver-state and trigger logs that \MUSRepair produces at each retry. We do not have equivalent per-turn logging for the non-repair methods, which limits fully symmetric cross-method comparison of failure channels. Post-hoc instrumentation of non-repair traces with the same solver checks would enable symmetric decomposition. We leave this to follow-up work. The benchmark covers three solver-structured domains, and whether the drift-dominance finding transfers to open-domain dialogue, where constraints are implicit and verification is harder, remains an open question. Finally, we evaluate a single repair routing design without ablating over trigger definitions, retry budgets, or alternative repair controllers. Different routing policies might shift the balance between contradiction and drift in the residual error distribution.
\vspace{-0.5em}

%=============================================================================
% 9  CONCLUSION
%=============================================================================
\section{Conclusion}

This paper introduced a solver-instrumented multi-turn benchmark that cleanly separates two failure modes, contradiction and satisfiable drift, and used it to evaluate four reasoning methods across four open-weight models. MUS-Repair produces significant gains in every setting after false-discovery correction, but the errors that survive are overwhelmingly drift. Models rarely contradict themselves after structured feedback, but they still forget prior commitments. This forgetting compounds with conversational depth, and accuracy declines steeply even on the strongest model, suggesting that long-horizon state maintenance remains an open challenge regardless of scale.

These findings point to a concrete gap in current evaluation practice. Solver-level contradiction checks are necessary but insufficient. Reliable multi-turn systems must also validate that the returned assignment respects the maintained state. Reporting contradiction and drift as separate channels, rather than merging them into a single accuracy number, exposes where the real residual risk lies.

\label{last-main-page}
\bibliography{refs}
\bibliographystyle{iclr2026/iclr2026_conference}

\clearpage
\appendix
\input{appendix}

\end{document}

%% file: iclr2026/math_commands.tex
\usepackage{amsmath,amsfonts,bm}

\def\eqref#1{equation~\ref{#1}}
\def\1{\bm{1}}

\DeclareMathAlphabet{\mathsfit}{\encodingdefault}{\sfdefault}{m}{sl}
\SetMathAlphabet{\mathsfit}{bold}{\encodingdefault}{\sfdefault}{bx}{n}

%% file: figures/method_comparison_v2.tex
% Method comparison figure (v2): moderate widths that don't force mid-word breaks,
% then \resizebox in the including figure to land on \textwidth.
\begin{tikzpicture}[
    node distance=0.45cm and 0.0cm,
    base_style/.style={
        rectangle,
        rounded corners,
        minimum width=#1,
        minimum height=.5cm,
        align=center,
        draw=black,
        line width=.6pt,
        font=\small,
        fill=iclrblue!8,
        text width=#1
    },
    arrow_style/.style={->, line width=0.8pt, draw=black!70},
    benefit_box/.style={
        draw=black!40,
        rounded corners,
        fill=white,
        text width=3.2cm,
        inner sep=0.2cm,
        font=\small
    },
    sig_panel/.style={
        draw=black!25,
        rounded corners,
        fill=black!2,
        inner sep=0.25cm
    }
    ]

    % ================================================================
    % LEFT PANEL: scheduling context header + 4-turn trajectory
    % ================================================================
    \node (Qhdr) [draw=black!15, rounded corners, fill=black!2,
                  minimum width=4.0cm, inner sep=0.08cm, align=center,
                  font=\small\ttfamily, text=black!75]
        {Scheduling: Mon/Tue/Wed\\to Alice, Bob, Carol};

    \node (Q11) [base_style=4.0cm, below=0.25cm of Qhdr] {Turn 1: Alice on Monday};
    \node (Q21) [base_style=4.0cm, below=of Q11] {Turn 2: Bob with Alice};
    \node (Q31) [base_style=4.0cm, below=of Q21] {Turn 3: Carol on Tuesday};
    \node (Q41) [base_style=4.0cm, below=of Q31, fill=red!12]
                {Turn 4: Bob $\neq$ Tuesday};

    \draw[arrow_style] (Q11) -- (Q21);
    \draw[arrow_style] (Q21) -- (Q31);
    \draw[arrow_style] (Q31) -- (Q41);

    % ================================================================
    % CENTER PANEL: properties of existing methods vs. MUS-Repair
    % ================================================================
    \node[benefit_box, anchor=north west] at ($(Qhdr.north east)+(.35cm,0)$) (benefits) {
        \textbf{Existing Methods:}\\[1pt]
        \xmark~No state tracking\\
        \xmark~No SAT verification\\
        \xmark~No drift detection\\[0.5em]
        \textbf{\MUSRepair:}\\[1pt]
        \cmark~Constraint ledger $L_t$\\
        \cmark~Z3 SAT check\\
        \cmark~MUS-guided repair
    };

    % ================================================================
    % RIGHT PANEL: formal method signatures, enclosed
    % ================================================================
    \node[sig_panel, anchor=north west, text width=6.5cm]
        at ($(benefits.north east)+(0.35cm,0)$) (sigpanel) {%
        \small
        \begin{tabular}{@{}p{5.5cm}@{\hspace{0.15cm}}p{0.4cm}@{}}
        \textbf{\Direct}\newline
            $G: \textcolor{orange}{H_{1:t}} \times u_t \rightarrow A_t$
            & \xmark \\[3pt]
        \textbf{\CoT}\newline
            $G: \textcolor{orange}{H_{1:t}} \times u_t \rightarrow \{r_k\}_{k=1}^{K}\!,\, A_t$
            & \xmark \\[3pt]
        \textbf{\LedgerMethod}\newline
            $G: \textcolor{orange}{H_{1:t}} \times u_t \times \textcolor{teal}{L_{t-1}} \rightarrow A_t$\newline
            $L_t \!\leftarrow\! L_{t-1} \cup E(u_t, a_t)$
            & \xmark \\[3pt]
        \textbf{\MUSRepair}\newline
            $G: \textcolor{orange}{H_{1:t}} \times u_t \times \textcolor{teal}{L_{t-1}} \rightarrow A_t$\newline
            $V: \textcolor{teal}{\mathrm{SAT}(L_t)} \wedge \mathrm{Satisfies}(A_t, L_t)$\newline
            $R: (\mathcal{T}_t, \textcolor{teal}{\mathcal{U}_t}) \rightarrow A'_t$
            & \cmark \\
        \end{tabular}
    };

\end{tikzpicture}

%% file: tables/benchmark_stats.tex
\begin{tabular}{llrrrr}
\toprule
Domain & Split & Turns [min,max] & Ent. & Vocab & Final \\
\midrule
Logic-Grid & 272/68/340 & 6.89 [4,10] & 4.00 & 4 & 11.57 \\
Scheduling & 272/68/340 & 7.06 [4,10] & 5.92 & 6 & 12.83 \\
Seating & 272/68/340 & 6.97 [4,10] & 7.01 & 7 & 11.20 \\
\bottomrule
\end{tabular}

%% file: tables/inferential_tests.tex
\begin{tabular*}{\linewidth}{@{\extracolsep{\fill}}l l S[table-format=2.2] S[table-format=+2.2] l S[table-format=1.4] S[table-format=2.1]@{}}
\toprule
Model & Method & {Acc.\,(\%)} & {$\Delta$ (pp)} & {95\% CI (pp)} & {$q_\mathrm{FDR}$} & {Retain\,\%} \\
\midrule
Qwen3-8B  & Direct (baseline)  & 28.19 & {---} & {---} & {---} & 5.1 \\
Qwen3-8B  & Chain-of-Thought & 27.91 & -0.19 & [$-$2.01, $+$1.63] & 0.8422 & 6.7 \\
Qwen3-8B  & Ledger           & 25.23 & -3.14 & [$-$4.99, $-$1.31] & 0.0034 & 5.7 \\
\rowcolor{iclrblue!8}
Qwen3-8B  & MUS-Repair       & \bfseries 30.01 & +2.03 & [$+$0.32, $+$3.71] & 0.0295 & \bfseries 8.8 \\
\addlinespace
Qwen3-32B & Direct (baseline)  & 28.93 & {---} & {---} & {---} & 6.5 \\
Qwen3-32B & Chain-of-Thought & 31.44 & +2.54 & [$+$0.73, $+$4.38] & 0.0187 & 2.5 \\
Qwen3-32B & Ledger           & 31.44 & +1.63 & [$-$0.93, $+$4.20] & 0.2204 & 23.6 \\
\rowcolor{iclrblue!8}
Qwen3-32B & MUS-Repair       & \bfseries 38.22 & +9.03 & [$+$6.98, $+$11.00] & 0.0002 & \bfseries 15.7 \\
\addlinespace
gpt-oss-20b  & Direct (baseline)  & 51.80 & {---} & {---} & {---} & 23.1 \\
gpt-oss-20b  & Chain-of-Thought & 50.35 & -1.39 & [$-$3.05, $+$0.27] & 0.1183 & 19.4 \\
gpt-oss-20b  & Ledger           & 53.70 & +1.91 & [$+$0.25, $+$3.60] & 0.0327 & 29.1 \\
\rowcolor{iclrblue!8}
gpt-oss-20b  & MUS-Repair       & \bfseries 68.71 & +16.20 & [$+$14.52, $+$17.90] & 0.0002 & \bfseries 48.3 \\
\addlinespace
gpt-oss-120b & Direct (baseline)  & 52.12 & {---} & {---} & {---} & 30.2 \\
gpt-oss-120b & Chain-of-Thought & 53.95 & +2.04 & [$+$0.36, $+$3.70] & 0.0295 & 24.2 \\
gpt-oss-120b & Ledger           & 50.02 & -2.29 & [$-$4.03, $-$0.59] & 0.0220 & 36.4 \\
\rowcolor{iclrblue!8}
gpt-oss-120b & MUS-Repair       & \bfseries 62.68 & +10.05 & [$+$8.40, $+$11.72] & 0.0002 & \bfseries 42.9 \\
\bottomrule
\end{tabular*}

%% file: tables/sat_but_wrong.tex
\begin{tabular}{@{}l S[table-format=4.0] S[table-format=2.0] S[table-format=3.0] >{\columncolor{iclrblue!8}}S[table-format=3.1] S[table-format=2.1] S[table-format=3.1]@{}}
\toprule
Model & {Drift} & {UNSAT} & {Other} & {Drift (\%)} & {UNSAT (\%)} & {Other (\%)} \\
\midrule
Qwen3-8B & 3970 & 0 & 0 & \bfseries 100.0 & 0.0 & 0.0 \\
Qwen3-32B & 3438 & 66 & 0 & \bfseries 98.1 & 1.9 & 0.0 \\
gpt-oss-20b & 1774 & 1 & 0 & \bfseries 99.9 & 0.1 & 0.0 \\
gpt-oss-120b & 2115 & 2 & 0 & \bfseries 99.9 & 0.1 & 0.0 \\
\bottomrule
\end{tabular}

%% file: tables/trigger_breakdown.tex
\begin{tabular}{@{}l S[table-format=5.0] S[table-format=5.0] S[table-format=5.0] S[table-format=5.0]@{}}
\toprule
Trigger & {Qwen3-8B} & {Qwen3-32B} & {gpt-oss-20b} & {gpt-oss-120b} \\
\midrule
Answer-Ledger Conflict        & 12089 & 10489 & 5300 & 6255 \\
Incomplete Assignment         &   218 &   528 & 1036 &   31 \\
Answer Parse Failure          &     0 &     2 &  664 &  810 \\
Constraint Extraction Failure &     0 &     2 &  559 &   36 \\
Unsatisfiable Ledger          &     1 &   212 &    2 &    4 \\
\bottomrule
\end{tabular}

%% file: appendix.tex
\section{Prompts and Message Schemas}
\label{sec:app-prompts}
Subsections A.1, A.2, and A.6 give the verbatim prompt strings. Subsections A.3, A.4, and A.5 describe the schema of assembled messages whose contents vary by domain, turn, and ledger state; we document the field structure rather than literal text.

\tcbset{
    promptstyle/.style={
        enhanced,
        breakable,
        boxrule=0.6pt,
        arc=1.0mm,
        left=1.8mm,
        right=1.8mm,
        top=1.4mm,
        bottom=1.4mm,
        fonttitle=\bfseries\footnotesize,
        coltitle=black,
        fontupper=\small,
        colback=white,
        colframe=iclrblue!40,
        colbacktitle=iclrblue!12,
        width=\linewidth,
        before skip=6pt,
        after skip=6pt
    }
}

\subsection{Main-Turn System Prompts}
\label{sec:app-main-system-prompts}
\begin{tcolorbox}[promptstyle,title={System Prompt \texttt{direct}}]
You solve multi-turn logical constraint satisfaction problems. Track all prior commitments and keep the final assignment consistent with every active constraint. Return only the final JSON solution.
\end{tcolorbox}
\begin{tcolorbox}[promptstyle,title={System Prompt \texttt{cot}}]
You solve multi-turn logical constraint satisfaction problems. Track all prior commitments and reason briefly before answering. Output at most 3 short bullets, then the final JSON solution.
\end{tcolorbox}
\begin{tcolorbox}[promptstyle,title={System Prompt \texttt{ledger\_only}}]
You solve a multi-turn logical constraint problem with an explicit ledger. Treat the ledger as committed state and keep your answer consistent with it and the new turn constraints. Return only JSON.
\end{tcolorbox}
\begin{tcolorbox}[promptstyle,title={System Prompt \texttt{mus\_repair}}]
You solve a multi-turn logical constraint problem with formal consistency checks and targeted repair signals. Use the ledger as committed state, address any repair signal directly, and provide a complete consistent solution. Return only JSON.
\end{tcolorbox}

\subsection{Additional System Prompts for Extraction and Retries}
\label{sec:app-extra-system-prompts}
\begin{tcolorbox}[promptstyle,title={Answer Retry System Prompt}]
You are a strict JSON formatter. Output one valid JSON object only. No markdown or prose.
\end{tcolorbox}
\begin{tcolorbox}[promptstyle,title={Constraint Extraction System Prompt}]
You extract formal constraints from an assistant answer. Extract only constraints introduced in the latest user turn. Do not restate full solution assignments unless they directly encode one allowed constraint.
\end{tcolorbox}

\subsection{Main User Message Schema}
\label{sec:app-user-assembly}
\begin{tcolorbox}[promptstyle,title={User Message Schema}]
\footnotesize
\textbf{Problem setup (first turn only):} include \texttt{Domain: <domain>} and \texttt{Entities: <entity list>} together with domain-specific context such as seat labels, time slots, or logic-grid categories.\\
\textbf{State block (ledger methods only):} include \texttt{Current ledger: <serialized constraints>}.\\
\textbf{Current turn block:} include \texttt{New constraints from user: <latest user message>}, optional \texttt{Repair signal: <trigger codes + optional MUS subset>}, and a domain-specific JSON schema hint.
\end{tcolorbox}

\subsection{Repair Signal Schema}
\label{sec:app-repair-signal}
\begin{tcolorbox}[promptstyle,title={Repair Signal Schema}]
\small
\textbf{REPAIR REQUIRED}
Detected issue classes are provided as \texttt{<trigger code> : <detail>}. An optional MUS subset is provided as \texttt{<constraint id> : "<constraint text>" (turn <t>)}. The required action is to return a revised JSON solution that resolves all listed issues.
\end{tcolorbox}

\subsection{Constraint Extraction Schema}
\label{sec:app-extraction-template}
\begin{tcolorbox}[promptstyle,title={Constraint Extraction Schema}]
\footnotesize
\textbf{System instruction:} extract only constraints introduced in the latest user turn from the assistant answer.\\
\textbf{User block fields:} include \texttt{Domain}, \texttt{Source turn}, and \texttt{Entities}; include \texttt{Latest user message} and \texttt{Assistant response}; specify the domain-specific allowed constraint vocabulary; enforce strict rules on type names, argument order, and turn locality; and require JSON output with key \texttt{"constraints"}.
\end{tcolorbox}

\subsection{Retry Prompt Templates}
\label{sec:app-retry-templates}
\begin{tcolorbox}[promptstyle,title={Answer Reformat Retry Prompt}]
\textbf{System:} strict JSON formatter, one valid JSON object only.\\
\textbf{User:} reformat the response as JSON only; preserve required entities as keys and add no commentary.
\end{tcolorbox}
\begin{tcolorbox}[promptstyle,title={Truncation Retry Prompt}]
Previous answer was clipped by token limit. Retry with one compact JSON object only: no bullets, no prose, no analysis.
\end{tcolorbox}

\par\vspace{0.5\baselineskip}
\section{Supplementary Tables}
\label{sec:app-tables}

\begin{table}[!htbp]
    \centering
    \caption{Domain-conditioned turn-level accuracy.}
    \label{tab:app-domain-breakdown}
    \footnotesize
    \input{tables/domain_breakdown}
\end{table}

\begin{table}[!htbp]
    \centering
    \caption{Truncation robustness check.}
    \label{tab:app-trunc-robust}
    \footnotesize
    \input{tables/truncation_robustness}
\end{table}

\begin{table}[!htbp]
    \centering
    \caption{Paired MUS-Repair tests against the strongest non-MUS comparator per model.}
    \label{tab:app-inferential-best-nonmus}
    \footnotesize
    \input{tables/inferential_vs_best_nonmus}
\end{table}

\begin{table}[!htbp]
    \centering
    \caption{Post-repair outcomes by trigger code in MUS traces.}
    \label{tab:app-repair-success}
    \footnotesize
    \input{tables/repair_success_by_trigger}
\end{table}

\begin{table}[!htbp]
    \centering
    \caption{Pairwise overlap of MUS error rows across models.}
    \label{tab:app-cross-overlap}
    \footnotesize
    \input{tables/cross_model_error_overlap}
\end{table}

\FloatBarrier

\section{Example Transcripts}
\label{sec:app-transcripts}

Three Qwen3-8B transcripts, one per domain, comparing \Direct and \MUSRepair turn by turn. \cmark\ marks a satisfying assignment, \xmark\ a violation.

\subsection{Transcript A: Scheduling (\texttt{scheduling\_249})}
\label{sec:transcript-scheduling}
\textbf{Setup.} Six activities (Sync, Testing, Meeting, QA, Planning, Design) must be assigned start times and durations. Constraints accumulate over four turns.

\begin{table}[!htbp]
\centering
\caption{Turn-by-turn outputs for \texttt{scheduling\_249}. \Direct 1/4, \MUSRepair 4/4.}
\label{tab:transcript-sched}
\small
\setlength{\tabcolsep}{3pt}
\resizebox{\linewidth}{!}{%
\begin{tabular}{@{}c p{0.32\linewidth} p{0.25\linewidth} c p{0.25\linewidth} c@{}}
\toprule
\textbf{Turn} & \textbf{New constraints} & \textbf{\Direct answer} & & \textbf{\MUSRepair answer} & \\
\midrule
1 & QA must start between slots 1--2.
  & QA$\to$1 & \cmark & QA$\to$2 & \cmark \\[3pt]
2 & Testing $\neq$ Design (simult.);\newline QA duration\,=\,3;\newline Design$\to$slot 9.
  & QA dur\,=\,3, Design$\to$9,\newline but Testing$\to$5, Design$\to$9\newline {\scriptsize(Testing dur 2 $\Rightarrow$ 5--6)}
  & \xmark\textsuperscript{\dag}
  & QA dur\,=\,3, Design$\to$9,\newline Testing$\to$4
  & \cmark \\[3pt]
3 & Testing duration\,=\,3.
  & Testing dur\,=\,3,\newline Design$\to$9 dur\,=\,2,\newline Meeting$\to$9 dur\,=\,2
  & \xmark
  & Testing$\to$4 dur\,=\,3;\newline Design$\to$9; QA$\to$2 dur\,=\,3
  & \cmark \\[3pt]
4 & Testing$\to$slot 7;\newline Planning$\to$slot 5.
  & Testing$\to$7 dur\,=\,3,\newline Design$\to$9 dur\,=\,2\newline {\scriptsize(overlap 9--10)}
  & \xmark
  & Testing$\to$7 dur\,=\,3,\newline Planning$\to$5, Design$\to$9
  & \cmark \\
\bottomrule
\multicolumn{6}{@{}l}{\textsuperscript{\dag}\scriptsize \Direct assigns default duration\,=\,2 to several activities, violating the implicit single-slot default.}
\end{tabular}%
}
\end{table}

\subsection{Transcript B: Logic Grid (\texttt{logic\_grid\_021})}
\label{sec:transcript-logicgrid}
\textbf{Setup.} Four people (Blake, Drew, Avery, Finley) are each assigned a unique value in three categories: color (Red/Blue/Green/Yellow), pet (Cat/Dog/Bird/Fish), and profession (Doctor/Artist/Teacher/Chef).

\begin{table}[!htbp]
\centering
\caption{Turn-by-turn outputs for \texttt{logic\_grid\_021}. \Direct 0/5, \MUSRepair 5/5.}
\label{tab:transcript-logic}
\small
\setlength{\tabcolsep}{3pt}
\resizebox{\linewidth}{!}{%
\begin{tabular}{@{}c p{0.32\linewidth} p{0.25\linewidth} c p{0.25\linewidth} c@{}}
\toprule
\textbf{Turn} & \textbf{New constraints} & \textbf{\Direct answer} & & \textbf{\MUSRepair answer} & \\
\midrule
1 & Finley.pet $<$ Drew.pet;\newline Finley.pet $\neq$ Avery.pet.
  & Finley$\to$Bird, Drew$\to$Dog\newline {\scriptsize(Bird $<$ Dog: wrong order)}
  & \xmark
  & Finley$\to$Cat, Drew$\to$Dog
  & \cmark \\[3pt]
2 & Blake.color $<$ Finley.color.
  & {\scriptsize(identical to turn 1)}
  & \xmark
  & Blake$\to$Red, Finley$\to$Yellow
  & \cmark \\[3pt]
3 & Drew.pet $\neq$ Finley.pet;\newline Avery.pet $\neq$ Drew.pet;\newline Avery.prof $\neq$ Finley.prof.
  & {\scriptsize(identical to turn 1)}
  & \xmark
  & Drew$\to$Dog, Avery$\to$Bird,\newline Finley$\to$Cat
  & \cmark \\[3pt]
4 & Drew $\to$ Chef.
  & Drew$\to$Chef, Finley$\to$Chef\newline {\scriptsize(\textbf{duplicate}: two Chefs)}
  & \xmark
  & Drew$\to$Chef, Finley$\to$Artist
  & \cmark \\[3pt]
5 & Blake.color $\neq$ Drew.color;\newline Drew $\neq$ Bird.
  & {\scriptsize(same as turn 4, still\newline two Chefs)}
  & \xmark
  & Drew$\to$Dog {\scriptsize(not Bird)};\newline Blake$\to$Red, Drew$\to$Blue
  & \cmark \\
\bottomrule
\end{tabular}%
}
\end{table}

\clearpage
\subsection{Transcript C: Seating (\texttt{seating\_062})}
\label{sec:transcript-seating}
\textbf{Setup.} Seven people (Diana, Ruby, Tina, Noah, Charlie, Frank, Karen) sit around a round table with positions 1--7. Constraints include fixed positions, adjacency prohibitions, and separation requirements.

\begin{table}[!htbp]
\centering
\caption{Turn-by-turn outputs for \texttt{seating\_062}. \Direct 1/4, \MUSRepair 3/4.}
\label{tab:transcript-seating}
\small
\setlength{\tabcolsep}{3pt}
\resizebox{\linewidth}{!}{%
\begin{tabular}{@{}c p{0.32\linewidth} p{0.25\linewidth} c p{0.25\linewidth} c@{}}
\toprule
\textbf{Turn} & \textbf{New constraints} & \textbf{\Direct answer} & & \textbf{\MUSRepair answer} & \\
\midrule
1 & Karen$\to$pos.\,3;\newline Karen not adjacent to Ruby.
  & Karen$\to$3, Ruby$\to$5
  & \cmark
  & Karen$\to$3, Ruby$\to$5
  & \cmark \\[3pt]
2 & Charlie not adjacent to Frank.
  & Charlie$\to$6, Frank$\to$7\newline {\scriptsize(\textbf{adjacent}: violates new constraint)}
  & \xmark
  & Charlie$\to$7, Frank$\to$4\newline {\scriptsize(separated by 3 positions)}
  & \cmark \\[3pt]
3 & Karen--Noah $\geq$1 apart;\newline Tina--Frank $\geq$2 apart.
  & Frank$\to$8\newline {\scriptsize(\textbf{invalid}: only 7 seats)}
  & \xmark
  & Noah$\to$7, Tina$\to$2,\newline Frank$\to$6
  & \cmark \\[3pt]
4 & Frank not adj.\ Ruby;\newline Noah not adj.\ Charlie;\newline Diana$\to$pos.\,6.
  & Karen$\to$4\newline {\scriptsize(\textbf{drift}: violates turn-1\newline at\_position(Karen,3))}
  & \xmark
  & Diana$\to$6, but Ruby$\to$5\newline adj.\ Frank$\to$1 {\scriptsize(wraps)}
  & \xmark \\
\bottomrule
\end{tabular}%
}
\end{table}

%% file: tables/domain_breakdown.tex
\begin{tabular}{llS[table-format=2.1]S[table-format=2.1]S[table-format=2.1]S[table-format=2.1]}
\toprule
Model & Domain & {Direct (\%)} & {CoT (\%)} & {Ledger (\%)} & {MUS-Repair (\%)} \\
\midrule
Qwen3-8B & Logic-Grid & 39.7 & 37.5 & 36.2 & 43.1 \\
Qwen3-8B & Scheduling & 35.0 & 35.1 & 29.1 & 34.7 \\
Qwen3-8B & Seating & 9.8 & 11.0 & 10.3 & 12.1 \\
\midrule
Qwen3-32B & Logic-Grid & 35.6 & 38.6 & 24.0 & 43.9 \\
Qwen3-32B & Scheduling & 41.1 & 44.8 & 56.9 & 55.7 \\
Qwen3-32B & Seating & 9.9 & 10.7 & 13.0 & 14.8 \\
\midrule
gpt-oss-20b & Logic-Grid & 57.3 & 56.1 & 60.8 & 81.4 \\
gpt-oss-20b & Scheduling & 64.4 & 64.6 & 68.8 & 85.7 \\
gpt-oss-20b & Seating & 33.5 & 30.1 & 31.2 & 38.7 \\
\midrule
gpt-oss-120b & Logic-Grid & 53.6 & 56.2 & 49.1 & 64.2 \\
gpt-oss-120b & Scheduling & 73.7 & 72.3 & 72.8 & 87.2 \\
gpt-oss-120b & Seating & 28.7 & 33.1 & 27.9 & 36.3 \\
\bottomrule
\end{tabular}

%% file: tables/truncation_robustness.tex
\begin{tabular}{llS[table-format=2.2]S[table-format=2.1]S[table-format=2.1]}
\toprule
Model & Method & {Trunc. (\%)} & {Acc. All (\%)} & {Acc. Non-trunc. (\%)} \\
\midrule
gpt-oss-120b & Direct & 0.37 & 52.1 & 52.3 \\
gpt-oss-120b & Chain-of-Thought & 0.23 & 53.9 & 54.1 \\
gpt-oss-120b & Ledger & 0.14 & 50.0 & 50.1 \\
gpt-oss-120b & MUS-Repair & 0.11 & 62.7 & 62.7 \\
\addlinespace
gpt-oss-20b & Direct & 1.75 & 51.8 & 52.7 \\
gpt-oss-20b & Chain-of-Thought & 1.53 & 50.4 & 51.1 \\
gpt-oss-20b & Ledger & 1.06 & 53.7 & 54.3 \\
gpt-oss-20b & MUS-Repair & 0.83 & 68.7 & 69.3 \\
\addlinespace
Qwen3-32B & Direct & 0.00 & 28.9 & 28.9 \\
Qwen3-32B & Chain-of-Thought & 0.02 & 31.4 & 31.4 \\
Qwen3-32B & Ledger & 0.00 & 31.4 & 31.4 \\
Qwen3-32B & MUS-Repair & 0.00 & 38.2 & 38.2 \\
\addlinespace
Qwen3-8B & Direct & 0.00 & 28.2 & 28.2 \\
Qwen3-8B & Chain-of-Thought & 0.00 & 27.9 & 27.9 \\
Qwen3-8B & Ledger & 0.00 & 25.2 & 25.2 \\
Qwen3-8B & MUS-Repair & 0.00 & 30.0 & 30.0 \\
\bottomrule
\end{tabular}

%% file: tables/inferential_vs_best_nonmus.tex
\begin{tabular}{llS[table-format=+2.2]lrr}
\toprule
Model & Comparator & {$\Delta$ Acc.\ (pp)} & 95\% CI (pp) & $p$ & $q_\mathrm{FDR}$ \\
\midrule
gpt-oss-120b & Chain-of-Thought & +8.01 & [$+$6.19, $+$9.83] & {$<$0.0001} & {$<$0.0001} \\
gpt-oss-20b & Ledger & +14.29 & [$+$12.70, $+$15.93] & {$<$0.0001} & {$<$0.0001} \\
Qwen3-32B & Chain-of-Thought & +6.49 & [$+$4.34, $+$8.66] & {$<$0.0001} & {$<$0.0001} \\
Qwen3-8B & Direct & +2.03 & [$+$0.34, $+$3.76] & 0.0178 & 0.0178 \\
\bottomrule
\end{tabular}

%% file: tables/repair_success_by_trigger.tex
\begin{tabular}{llS[table-format=4.0]S[table-format=2.1]S[table-format=3.1]}
\toprule
Model & Trigger & {Rows} & {Repair Acc. (\%)} & {Repair SAT (\%)} \\
\midrule
Qwen3-8B & Answer-Ledger Conflict & 4131 & 4.0 & 100.0 \\
Qwen3-8B & Incomplete Assignment & 139 & 65.5 & 100.0 \\
Qwen3-8B & Unsatisfiable Ledger & 1 & 0.0 & 100.0 \\
\addlinespace
Qwen3-32B & Answer-Ledger Conflict & 3643 & 5.5 & 99.6 \\
Qwen3-32B & Incomplete Assignment & 275 & 69.8 & 100.0 \\
Qwen3-32B & Unsatisfiable Ledger & 87 & 6.9 & 17.2 \\
Qwen3-32B & Constraint Extraction Failure & 2 & 0.0 & 100.0 \\
\addlinespace
gpt-oss-20b & Answer-Ledger Conflict & 2402 & 33.3 & 100.0 \\
gpt-oss-20b & Incomplete Assignment & 735 & 33.3 & 99.9 \\
gpt-oss-20b & Answer Parse Failure & 449 & 23.4 & 100.0 \\
gpt-oss-20b & Constraint Extraction Failure & 429 & 22.6 & 100.0 \\
\addlinespace
gpt-oss-120b & Answer-Ledger Conflict & 2471 & 21.2 & 99.9 \\
gpt-oss-120b & Answer Parse Failure & 416 & 18.5 & 99.8 \\
gpt-oss-120b & Constraint Extraction Failure & 32 & 12.5 & 96.9 \\
gpt-oss-120b & Incomplete Assignment & 30 & 26.7 & 100.0 \\
\bottomrule
\end{tabular}

%% file: tables/cross_model_error_overlap.tex
\begin{tabular}{llS[table-format=4.0]S[table-format=1.3]S[table-format=1.3]S[table-format=1.3]}
\toprule
Model A & Model B & {Overlap} & {Jaccard} & {Share A} & {Share B} \\
\midrule
Qwen3-8B & Qwen3-32B & 3143 & 0.726 & 0.792 & 0.897 \\
Qwen3-8B & gpt-oss-20b & 1687 & 0.416 & 0.425 & 0.950 \\
Qwen3-8B & gpt-oss-120b & 1945 & 0.470 & 0.490 & 0.919 \\
Qwen3-32B & gpt-oss-20b & 1638 & 0.450 & 0.467 & 0.923 \\
Qwen3-32B & gpt-oss-120b & 1892 & 0.507 & 0.540 & 0.894 \\
gpt-oss-20b & gpt-oss-120b & 1412 & 0.569 & 0.795 & 0.667 \\
\bottomrule
\end{tabular}